\documentclass[10pt,twocolumn,letterpaper]{article}

\usepackage{iccv}              
\usepackage{multirow}
\usepackage{booktabs}
\usepackage{colortbl}
\usepackage{adjustbox}
\definecolor{tabfirst}{rgb}{1, 0.7, 0.7}   
\definecolor{tabsecond}{rgb}{1, 0.85, 0.7}  
\definecolor{tabthird}{rgb}{1, 1, 0.7}

\newcommand{\best}[1]{\cellcolor{tabfirst}#1}
\newcommand{\second}[1]{\cellcolor{tabsecond}#1}
\newcommand{\third}[1]{\cellcolor{tabthird}#1}
\newcommand{\mystrut}{\rule[-0.4ex]{0pt}{1.7ex}}
\newcommand{\highlight}[2]{\colorbox{#1}{\mystrut#2}}
\definecolor{iccvblue}{rgb}{0.21,0.49,0.74}
\usepackage[pagebackref,breaklinks,colorlinks,allcolors=iccvblue]{hyperref}

\def\paperID{3312} 
\def\confName{ICCV}
\def\confYear{2025}

\title{ETA: Energy-based Test-time Adaptation for Depth Completion}

\author{
Younjoon Chung\thanks{Equal contribution} \quad
Hyoungseob Park\footnotemark[1] \quad
Patrick Rim\footnotemark[1] \quad
Xiaoran Zhang\quad
Jihe He\\
Ziyao Zeng\quad
Safa Cicek$^{\dagger}$\quad
Byung-Woo Hong$^{\ddagger}$\quad
James S. Duncan\quad
Alex Wong\\[0.2cm]
Yale University\quad
$^{\dagger}$UCLA\quad
$^{\ddagger}$Chung-Ang University\\[0.1cm]
{\tt\small \{younjoon.chung, hyoungseob.park, patrick.rim, xiaoran.zhang\}@yale.edu}\\
{\tt\small  safacicek@ucla.edu,\quad hong@cau.ac.kr, \{james.duncan, alex.wong\}@yale.edu }\\
}

\begin{document}
\maketitle

\begin{abstract}
We propose a method for test-time adaptation of pretrained depth completion models. Depth completion models, trained on some ``source'' data, often predict erroneous outputs when transferred to ``target'' data captured in novel environmental conditions due to a covariate shift. The crux of our method lies in quantifying the likelihood of depth predictions belonging to the source data distribution. The challenge is in the lack of access to out-of-distribution (target) data prior to deployment.
Hence, rather than making assumptions regarding the target distribution, we utilize adversarial perturbations as a mechanism to explore the data space.
This enables us to train an energy model that scores local regions of depth predictions as in- or out-of-distribution.
We update the parameters of pretrained depth completion models at test time to minimize energy, effectively aligning test-time predictions to those of the source distribution. We call our method ``Energy-based Test-time Adaptation'', or ETA for short.
We evaluate our method across three indoor and three outdoor datasets, where ETA improve over the previous state-of-the-art method by an average of 6.94\% for outdoors and 10.23\% for indoors.
Project Page: \href{https://fuzzythecat.github.io/eta}{https://fuzzythecat.github.io/eta}.
\end{abstract}

\section{Introduction}
\label{sec:intro}




Reconstructing the 3-dimensional (3D) environment is essential for spatial tasks such as autonomous driving, physical AI agents, and extended reality. Many of these tasks rely on sensor platforms equipped with visual (e.g., camera) and range (e.g., LiDAR or radar) sensors. These sensors are complementary in that: Cameras capture RGB images with (dense) irradiance at each pixel, but 3D reconstruction from visual data is ill-posed due to the loss of a dimension in perspective projection. Range sensors measure coordinates or point clouds of the 3D environment or scene; likewise these point clouds can be estimated from images via Structure-from-Motion (SfM) or Visual Inertial Odometry (VIO). But in both cases, the point clouds tend to be sparse. Depth completion is the multimodal 3D reconstruction task of estimating a dense depth map from an RGB image and its synchronized sparse point cloud, often projected onto the image plane as a sparse depth map.

As depth completion models, pretrained on ``source'' datasets, are deployed to support a range of spatial tasks, they often encounter novel environments or 3D scenes and conditions. Hence, there exists a covariate shift when transferring a depth completion model to ``target'' testing data. Such distributional shifts arise from nuisance variables, e.g., changes in illumination, occlusion, that cause changes in object appearance,  which lead to model degradations.

Mitigating this covariate shift can be achieved by means of domain adaptation \cite{farahani2021brief}, which updates the model parameters to fit the target testing 
data, but often requires access to  source and target datasets with ground truth. Source-free \cite{li2024comprehensive} and unsupervised \cite{wilson2020survey} domain adaptation relax the assumption of having access to source data and ground truth, respectively. However, they still require access to the testing dataset and multiple passes through it for model updates. These requirements pose a hurdle for adoption in real-time spatial applications. Instead, we consider the problem of test-time adaptation (TTA), where one is given access to the testing data in a stream, i.e., one batch at a time, without access to ground truth. Hence, TTA methods must update the parameters of a pretrained model in an unsupervised manner, under low-computational budget, with time constraints, and without the ability to revisit previously-seen examples. 

To this end, we aim to model the shift or domain gap between the source and target distributions in order to guide test-time updates to a pretrained model to yield predictions resembling those of the source distribution. This will be facilitated by implicitly modeling the data distributions through an energy function. The energy function will quantify the likelihood of the data belonging the source distribution by associating a scalar (energy) to the data, where low energy is assigned to more probable (source-like, in-distribution) data points and high energy to out-of-distribution data points. A key insight is that (sparse) point clouds or depth estimates are lower dimensional and tend to exhibit a smaller covariate shift than RGB images  \cite{park2024test}. Hence, rather than quantifying the energy of the inputs, we will define an energy function over depth predictions, conditioned on its sparse depth map, allowing an energy model to validate the likelihood of predictions against measured coordinates of the 3D environment. Further, we estimate energy on a patch- or region-level to localize out-of-distribution predictions, which enables targeted updates.   

However, it remains a challenge to determine what constitutes out-of-distribution data points without making assumptions on the target test-time data to be observed in the future. Instead, we posit that adversarial perturbations of the input RGB image and sparse depth map can serve as a means of exploring the data space of predicted depth maps. Thus, we train an energy model using source dataset and its out-of-distribution examples, synthesized by adversarial perturbations, in a preparation stage. Once trained, the energy model will be frozen and deployed with the pretrained depth completion model to the testing environment. For each new example (or batch) observed at test time, we will update the parameters of the depth completion model to minimize energy, which effectively aligns predictions (and the model) back to those of the source distribution. We term our method ``Energy-based Test-time Adaptation'', or ETA for short. We extensively evaluate ETA across three indoor and three outdoor datasets for three distinct adaptation scenarios, involving indoor-to-indoor, outdoor-to-outdoor, and outdoor-to-indoor, where we consistently outperform baseline methods and achieve the state of the art.

\textbf{Our contributions:}
(1) We propose a novel test-time adaptation framework, ETA, that utilizes an energy model to quantify the likelihood of depth predictions belonging to the source distribution and adapts a pretrained model to the target distribution by minimizing the energy in predictions. 
(2) We utilize adversarial perturbations as a mechanism to explore the data space, which mitigates the challenge in the lack of out-of-distribution data for training the energy model.
(3) We propose a patch-based energy model that localizes high energy regions, allowing further targeted updates to specific regions within the depth predictions.

\section{Related Works}
\label{sec:relatedwork}

\noindent\textbf{Depth completion} is the multimodal 3D reconstruction task of inferring a dense depth map from a single image and its associated sparse depth map, unlike single image or monocular depth estimation \cite{eigen2014depth,godard2019digging, lao2024depth,saxena2005learning,upadhyay2023enhancing,wong2019bilateral,zeng2024rsa,zeng2024priordiffusion,zeng2024wordepth,gangopadhyay2025extending}. The sparse depth map is the projection of a point cloud measured by LiDAR \cite{xia2023quadric,xie2023sparsefusion}, radar \cite{rim2025radar,singh2023depth}, or visual inertial odometry systems \cite{fei2019geo,wong2020unsupervised}. Depth completion can be trained in a supervised \cite{cheng2020cspn++,ezhov2024all,jaritz2018sparse,park2020non,uhrig2017sparsity,yang2019dense} or unsupervised \cite{chancan20253d,yan2023desnet,lopez2020project,ma2019self,liu2022monitored,shivakumar2019dfusenet,wong2020unsupervised,wong2021adaptive,wong2021learning,wong2021unsupervised,wu2024augundo,rim2025protodepth,chen2024uncle} manner. We focus on supervised methods, where \cite{jaritz2018sparse, ma2019self, yang2019dense} introduce strategies to fuse image and depth features at both early and late network stages. In contrast, \cite{hu2021penet} advocates maintaining separate networks for each modality. To enhance the depth encoder with image cues, \cite{li2020multi} proposes a multi-scale cascaded hourglass design, and \cite{cheng2020cspn++} leverages convolutional spatial propagation layers; this is extended by \cite{park2020non} to non-local propagation for refining depth estimates based on confidence and learnable affinity, and by \cite{lin2022dynamic} to dynamically adaptive spatial propagation. Various studies also estimate depth uncertainty: \cite{eldesokey2020uncertainty,qu2021bayesian,qu2020depth} model prediction variance, while \cite{van2019sparse} applies confidence maps to merge multiple depth predictions. Further, \cite{qiu2019deeplidar, xu2019depth, zhang2018deep} utilize surface normals as an auxiliary cue for depth refinement. \cite{kam2022costdcnet} incorporates cost volumes. \cite{rho2022guideformer} introduces transformer-based cross-modal attention, whereas \cite{youmin2023completionformer} combines both convolutional and transformer operations. More recently, \cite{tang2024bpnet} presents a multi-scale approach with bilateral propagation and fusion of depth and image. Although these methods achieve high performance in controlled environments and on public benchmarks, distribution shifts arising from changes in illumination, sensor noise, or scene types can degrade performance, calling for robust adaptation strategies to new domains.

\vspace{1pt}\noindent\textbf{Domain Adaptation} (DA) \cite{farahani2021brief} aims to bridge the gap between data distributions in training (source) and testing (target). Unsupervised domain adaptation (UDA) \cite{wilson2020survey,ganin2015unsupervised,peng2019moment,cicek2019unsupervised} typically assumes access to both source and target data, which frequently does not hold in practical depth completion scenarios. More recently, \cite{kim2021domain} introduces the source-free DA paradigm that rely solely on target data for adaptation \cite{li2024comprehensive}, but requires multiple passes over the target.
While these domain adaptation strategies have shown potential in tasks like classification or semantic segmentation \cite{li2020model, shin2022mm}, few approaches specifically address depth completion \cite{lopez2020project}, which often has to cope with not only domain shifts in image appearance, but also in sparse depth distribution (e.g.,, density changes in LiDAR points). Moreover, conventional DA methods typically require computationally expensive retraining or fine-tuning phases. Such constraints motivate test-time adaptation methods for depth completion that can adapt efficiently on-the-fly.

\noindent\textbf{Test-Time Adaptation} (TTA) \cite{liang2024survey} focuses on adapting models to new input distributions at test time, without access to the original source (training) data at test time. Unlike UDA, TTA assumes a single pass over the target dataset, making it well-suited for real-time robotic perception or autonomous driving. Early TTA works \cite{sun_test-time_2020,liu_ttt_2021,lao2024sub} used auxiliary self-supervised tasks (e.g., rotation prediction) to guide adaptation. More recent methods update statistics in batch normalization layers \cite{schneider_improving_2020}, minimize entropy \cite{wang2021tent} for classification tasks. Follow-ups include strategies for stable parameter updates \cite{niu_efficient_2022, zhang_memo_2022}, leveraging pseudo-label refinement \cite{wang2022continual} or contrastive objectives \cite{chen2022contrastive}. The work most closely related to ours are concurrent work \cite{zhang2025progressive} and TEA \cite{yuan2024tea}, which transforms pretrained classifiers into energy-based models to learn the underlying target distribution. However, most TTA approaches emphasize classification or segmentation; only a few extend to depth estimation or  \textit{completion} \cite{yeo2023rapid, park2024test}, which deals with a continuous regression objective. A recent direction uses depth priors \cite{ke2024repurposing} derived from foundation models \cite{rombach2022foundation} for zero-shot generalization across different domains; however, while these generative methods achieve notable gains, they typically require non-trivial computational overhead at test time. Our approach offers a complementary alternative, focusing on a direct energy-based criterion that effectively bridges the gap across different domains for \textit{depth completion}, while maintaining runtime efficiency.

\noindent\textbf{Energy-Based Models} (EBMs) are probabilistic models that define a probability distribution implicitly through an energy function, where lower value corresponds to higher likelihood. Unlike conventional probabilistic approaches, EBMs do not require an explicit computation of the normalization constant. Thus, density estimation with EBMs can be viewed as a nonlinear regression problem, allowing considerable flexibility in choosing any suitable nonlinear function as the energy function. Recent studies have demonstrated that combining EBMs with neural networks enables accurate modeling of complex data distributions, achieving significant success in various tasks such as image generation \cite{ngiam2011energy, oord2016pixel, du2019generation}, discriminative learning \cite{fredrik2020ebm}, and density estimation \cite{vincent2011connection, song2019sliced}. Despite these advancements, the use of EBMs for TTA remains relatively unexplored \cite{yuan2024tea,zhang2025progressive}, particularly in the context of depth completion, where no prior work currently exists. In this work, we propose a novel energy-based approach to explicitly bridge domain gaps by minimizing the estimated energy associated with erroneous predictions during TTA for depth completion.

\begin{figure*}
    \centering
    \includegraphics[width=1.0\linewidth]{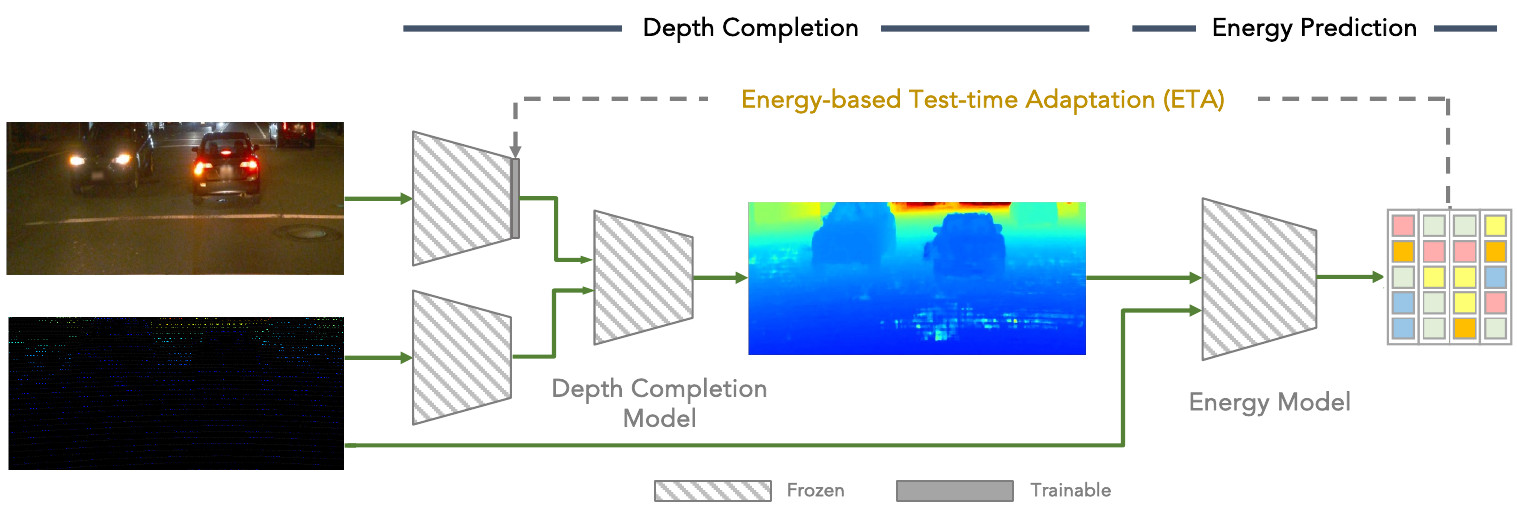}
    \vspace{-7mm}
    \caption{\textit{Overview.}
    Our energy-based test-time adaptation framework. The depth completion model predicts a dense depth map from an RGB image and sparse depth inputs. The energy model estimates a region-wise energy map of the dense depth map, which indicates the likelihood of the prediction associated with source data distribution.
    The adaptation layer is placed within the RGB encoder of the depth completion model to adapt the RGB features to the unseen target domain.
    During adaptation, only the adaptation layer (and optionally BatchNorm) is updated, while all other parameters remain frozen. The adaptation is guided by conventional unsupervised losses and the proposed energy minimization, which updates the adaptation layer to generate plausible 3D scene geometry in the prediction that align with scenes observed in the source data. Solid arrows denote data flow, while dashed arrows represent adaptation steps.
    }
    \label{fig:pipeline}
    \vspace{-3mm}
\end{figure*}

\section{Method Formulation}
\label{sec:method_formulation}

\label{sec:preliminary}
\vspace{1pt}\noindent\textbf{Depth Completion Models.} Let $\Omega \subset \mathbb{R}^2$ denote the image space. An RGB image is defined as $I: \Omega \to \mathbb{R}^3$, a sparse depth map as $z: \Omega_z \subset \Omega \to \mathbb{R}_{+}$, and a dense depth map as $d: \Omega \to \mathbb{R}_+$. A depth completion model $f_\theta$ estimates a dense depth map from an RGB image and its corresponding sparse depth measurements. This is given by $\hat{d} = f_\theta(I, z)$, where $\hat{d}: \Omega \to \mathbb{R}_{+}$ is the estimated dense depth map.

We assume that a depth completion model $f_{\theta}$ is trained with supervision on a \textit{source} dataset $\mathcal{D}_s = \{(I_s^{(n)}, z_s^{(n)}, d_s^{(n)})\}_{n=1}^{N_s}$, where $d_s$ denotes the ground-truth depth map. The model will be deployed to a \textit{target} dataset $\mathcal{D}_t \;=\; \bigl\{\,(I_t^{(n)}, z_t^{(n)})\bigr\}_{n=1}^{N_t}$. We assume the underlying test-time data distribution is represented by $\mathcal{D}_t$ and is out-of-distribution with respect to  $\mathcal{D}_s$ due to \textit{covariate shifts}.

\vspace{1pt}\noindent\textbf{Energy-based Models.} Let $\mathcal{X}$ be the input (feature) space. Energy-based models (EBMs) define a function $E_\phi: \mathcal{X} \to \mathbb{R}$, which assigns a scalar value, \textit{energy}, to each input sample $x \in \mathcal{X}$. Lower energy indicates higher likelihood that a sample $x$ belongs to the distribution modeled by the EBM, which is represented by the Boltzmann distribution \cite{lifshitz1980statistical}:
\begin{equation}
\label{eq:alt}
    p_\phi(x) = \frac{\exp(-E_\phi(x))}{Z(\phi)},
\end{equation}
where $Z(\phi) = \int_\mathcal{X} \exp(-E_\phi(x))\mathrm{d}x$ is the partition function, is parameterized by weight parameters $\phi$.


\subsection{Energy-based Model Formulation}
\label{sec:energy_formulation}
\vspace{1pt}\noindent\textbf{Depth Energy-Based Models.} We define an energy function $E_\phi$, parameterized by $\phi$, to quantify the likelihood of the depth predictions $\hat{d}$ belonging to the source distribution. As the depth completion is an ill-posed task, we select for one of the infinitely many 3D scenes by conditioning on the given sparse depth input $z$,
\begin{equation}
    E_\phi : (\hat{d}; z) \to e,\quad e:\Omega \to [0,1],
\end{equation}
where $e$ denotes energy.  
Instead of estimating a single energy value for the entire predicted depth map~\cite{yuan2024tea}, we propose a region-based approach, where the energy model produces a lower-resolution energy map $e$, and each element of $e$ corresponds to the likelihood of a local region in the full resolution depth map.
Lower energy values indicate that the predicted depth $\hat{d}$ is likely in-distribution (i.e., less erroneous), while higher energy values suggest potential errors (i.e., out-of-distribution predictions). 
This approach naturally imposes spatial regularity in the energy predicted, where we balance the ability to localize error but avoid spurious changes in the energy map. Additionally, by conditioning on the sparse depth map, we enable the energy model to validate the depth predictions against 3D coordinates measured within the environment, which aids in determining plausibility of the predicted 3D scene.

\vspace{1pt}\noindent\textbf{Mapping Distance to Energy.}  
To train the energy model $E_\phi$, we begin by modeling the target energy as proportional to the prediction error with respect to ground truth depth, for both the depth predictions $\hat{d_s}$ on the original source data and  $\tilde{d_s}$ on the (synthesized) out-of-distribution data.
For each patch $\Omega_p \subset \Omega$, we compute the mean squared error (MSE):
\vspace{-4mm}
\begin{equation}
    \Delta(\Omega_{p}) = \frac{1}{|\Omega_{p}|}\sum_{x \in \Omega_{p}}\bigl(\hat{d}_s(x)-{d}_s(x)\bigr)^2.
    \label{eq_distance}
\vspace{-1mm}
\end{equation}
for $\hat{d_s}$ and likewise for $\tilde{d_s}$.  We formulate the target energy $y$ as the complement of the Gibbs probabilistic distribution: 
\vspace{-1mm}
\begin{equation}
y = 1 - \exp\bigl(-\Delta/\tau\bigr),
\label{eq_gibbs_mapping}
\end{equation}
where $\tau$ is a temperature term.
Under this definition, small differences (minimal perturbations) yield energy values close to 0, while larger differences produce higher energy values near 1, marking unreliable or out-of-distribution predictions. This negative exponential mapping bounds energies within $[\hspace{0.05em}0,1]$, matching the output range of $E_\phi$. Note that the mapped energy is a distribution conditioned by depth completion parameters $\theta$. 

\vspace{1pt}\noindent\textbf{Generating out-of-distribution samples.} While the energy model $E_\phi$ must learn to distinguish between in-distribution from out-of-distribution depth predictions, we only have access to the source dataset $\mathcal{D}_s$ during training, or the preparation stage. Therefore, out-of-distribution examples need to be generated from source data. However, it is difficult to model test-time variations without making strong assumptions regarding the target distribution, which would limit the generalizability of the method. Instead, we repurpose \textit{adversarial perturbations} \cite{goodfellow2014explaining,wong2021stereopagnosia,berger2022stereoscopic} of the input RGB image $I_s$ and sparse depth map $z_s$ as a mechanism to explore the space of depth predictions $\hat d_s=f_\theta(I_s,z_s)$. This enables us to simulate the out-of-distribution examples that induce errors to the depth prediction. 
Adversarial perturbations adjust the inputs to statistical outliers in the blind high-density regions of the data space \cite{goodfellow2014explaining}, which will cause out-of-distribution errors. This encompasses a broad range of variations that covers more than a single ``target domain''. Hence, this enables the reuse of a single energy model across different testing dataset or distributions.

The adversarial perturbations correspond directly to the model's failure modes, which presents the energy model $E_\phi$ with out-of-distribution data that are assigned high energy as formulated by Eq. \ref{eq_gibbs_mapping}.
We choose Fast Gradient Sign Method (FGSM) \cite{goodfellow2014explaining,wong2021stereopagnosia} for perturbing the inputs $(I_s, z_s)$:
\begin{align}
\tilde{I_s} &= I_s + \epsilon_{I} \cdot \mathrm{sign}\bigl(\nabla_{I_s} \mathcal{L}_{\text{sup}} (\hat{d_s}, d_s)\bigr) \\
\tilde{z_s} &= z_s + \epsilon_{z} \cdot \mathrm{sign}\bigl(\nabla_{z_s} \mathcal{L}_{\text{sup}}(\hat{d_s}, d_s)\bigr)
\end{align}
where $\epsilon_I$ and $\epsilon_z$ are parameters controlling the perturbation magnitudes, and $\mathcal{L}_{\mathrm{sup}}$ is the supervised loss between the predicted depth $\hat{d_s}$ and ground-truth depth $d_s$. This is accomplished by a forward pass through the depth completion model $f_\theta$ with the original inputs $(I_s, z_s)$ and computing the gradient with respect to $I_s$ and $z_s$. The perturbations are constrained within an $l_\infty$ ball of radius $\epsilon$. After adding the perturbations to yield $(\tilde{I_s}, \tilde{z_s})$, we make a second forward pass to obtain the synthesized out-of-distribution example $\tilde{d}_s = f_\theta(\tilde{I_s}, \tilde{z_s})$, from which we compute the target energy $y$ (see  \cref{eq_distance,eq_gibbs_mapping}) for training our energy model. 


\vspace{1pt}\noindent\textbf{Training Energy-based Model.}
Since marginalizing over $Z(\phi)$ is intractible, we instead pose the learning of an energy model as a discriminative task.
Therefore, we optimize the energy model $E_\phi$ by minimizing cross entropy over all patches between the predicted energy map $e$ and the target energy map $y$, which enables us to model the probability of in- (low energy) and out-of-distribution (high energy) data:
\vspace{-0.3cm}
\begin{equation}
    \mathcal{L}_{\mathrm{energy}} = -\frac{1}{|\Omega_p|}\sum_{x \in \Omega_p} y(x) \log \big(y(x) / e(x)\big) 
\end{equation}


Once trained, the energy-based model will assign lower energy values to predictions aligned with the source data distribution and higher energy to erroneous or out-of-distribution predictions. Note: the energy model trained on one depth completion model is not applicable for other models since the trained distribution is conditioned by the depth completion model parameters.


\subsection{Test-Time Adaptation via Energy Guidance}
\label{sec:energy_tta}
The energy model $E_\phi$ is deployed in conjunction with the depth completion model $f_\theta$ to infer depth on the target testing data $\mathcal{D}_t$. Both the parameters of $E_\phi$ and $f_\theta$ are fixed or frozen.
Following \cite{park2024test}, we insert a lightweight adaptation module $m_\psi$ in the pretrained image encoder of $f_\theta$. Only the adaptation parameters $m_\psi$ and BatchNorm statistics of $f_\theta$ are updated at inference. Fig.~\ref{fig:pipeline} shows an overview.

\vspace{1pt}\noindent\textbf{Energy-based Alignment.} 
As high-energy test-time predictions $\hat{d}_t = f_{\theta,\psi}(I_t, z_t)$ correspond to out-of-distribution predictions, we posit that minimizing their energy will improve the fidelity of predictions by aligning them back to those of the source data distribution. Hence, minimizing the energy $e$ predicted from $E_\phi$ is analogous to minimizing the likelihood of error learned from the source data.
This motivates an energy-based adaptation loss that aims to reduce the energy of $\hat{d}_t(x)$ conditioned on $z_t(x)$:
\begin{equation}
\ell_{\mathrm{energy}} = -\frac{1}{|\Omega_p|} \sum_{x\in\Omega_p}\log\bigl(1 -  E_\phi(\hat{d}_t, z_t\bigr)(x)).
\end{equation}


\vspace{1pt}\noindent\textbf{Sparse Depth Consistency.} Sparse depth measurements provide partial yet accurate information about the 3D scene structure. To ensure the predicted depth map $\hat{d}_t$ aligns accurately with these sparse measurements, we minimize the $L_1$ distance between the sparse depth points $z_t$ and their corresponding predictions $\hat{d}_t$:
\begin{equation}
\label{eq:sparse}
    \ell_{\text{sparse}} = \frac{1}{|\Omega_z|}\sum_{x \in \Omega_z}|\hat{d}_t(x) - z_t(x)|,
\end{equation}
where $\Omega_z$ denotes the set of pixels with valid sparse depth measurements.

\vspace{1pt}\noindent\textbf{Local Smoothness.}  
Assuming that 3D scenes exhibit local smoothness and connectivity, we apply a similar constraint to the predicted depth map \( \hat{d}_t \). To achieve this, we use an \( L_1 \) penalty on the gradients of \( \hat{d}_t \) along the horizontal and vertical directions, represented by \( \partial_X \) and \( \partial_Y \). To maintain discontinuities at the object boundaries, we discount these penalties with spatially varying weights \( \lambda_X \) and \( \lambda_Y \), which are defined based on image gradients. Specifically, we set \( \lambda_X(x) = e^{-|\partial_X I_t(x)|} \) and \( \lambda_Y(x) = e^{-|\partial_Y I_t(x)|} \).
\begin{equation}
\label{eq:smooth}
    \resizebox{0.88\hsize}{!}{$
    \ell_{\text{smooth}} =\scalebox{1.5}{$\frac{1}{|\Omega|}$} \sum\limits_{x\in\Omega} \lambda_X(x) |\partial_X \hat{d}_t(x)| + \lambda_Y(x) |\partial_Y \hat{d}_t(x)|.
    $}
\end{equation}

\begin{table*}[t]
\scriptsize
\setlength\tabcolsep{1.5pt}
\renewcommand{\arraystretch}{1.1} 
\centering
\resizebox{1.00\textwidth}{!}{%
\begin{tabular}{cl cc cc cc c cc cc cc}
\toprule
 &  & \multicolumn{6}{c}{\textbf{Outdoor (KITTI)}} & \multicolumn{1}{c}{} & \multicolumn{6}{c}{\textbf{Indoor (VOID)}} \\
\cmidrule(lr){3-8} \cmidrule(lr){10-15}
Model & Method 
      & \multicolumn{2}{c}{VKITTI-FOG} 
      & \multicolumn{2}{c}{nuScenes} 
      & \multicolumn{2}{c}{Waymo} 
      & \multicolumn{1}{c}{} &
      \multicolumn{2}{c}{NYUv2} 
      & \multicolumn{2}{c}{SceneNet} 
      & \multicolumn{2}{c}{ScanNet} \\
\cmidrule(lr){3-4}\cmidrule(lr){5-6}\cmidrule(lr){7-8}\cmidrule(lr){10-11}\cmidrule(lr){12-13}\cmidrule(lr){14-15}
 &  & MAE$\downarrow$ & RMSE$\downarrow$ 
      & MAE$\downarrow$ & RMSE$\downarrow$ 
      & MAE$\downarrow$ & RMSE$\downarrow$
      &  & 
      MAE$\downarrow$ & RMSE$\downarrow$
      & MAE$\downarrow$ & RMSE$\downarrow$
      & MAE$\downarrow$ & RMSE$\downarrow$ \\
\midrule
\multirow{4}{*}{MSG-CHN \cite{li2020multi}} 
    & Pre-trained     
          & 2.842 & 6.557  
          & 3.331 & 6.449  
          & 1.107 & 2.962  
          & & 1.040 & 1.528  
          & 0.281 & 0.645  
          & 0.687 & 1.201 \\
    & CoTTA \cite{wang2022continual} 
          & \third{0.730} & \third{3.330}  
          & \third{3.157} & \third{6.434}  
          & \third{0.655} & \third{2.213}  
          & & \third{0.876} & \third{1.148}  
          & \third{0.223} & \third{0.498}  
          & \third{0.619} & \third{1.141} \\
    & ProxyTTA-fast \cite{park2024test} 
          & \second{0.728} & \second{3.087}  
          & \second{2.834} & \second{6.096}  
          & \second{0.608} & \second{1.921}  
          & & \second{0.699} & \second{1.120}  
          & \second{0.192} & \second{0.424}  
          & \second{0.302} & \second{0.480} \\
    & ETA (Ours)      
          & \best{0.703}  & \best{2.996}  
          & \best{2.666}  & \best{6.073}  
          & \best{0.583}  & \best{1.907}  
          & & \best{0.561}  & \best{0.850}  
          & \best{0.187}  & \best{0.401}  
          & \best{0.298}  & \best{0.460} \\
\midrule
\multirow{6}{*}{NLSPN \cite{park2020non}} 
    & Pre-trained     
          & 1.309 & 7.423  
          & 2.656 & \third{6.146}  
          & 1.175 & 3.078  
          & & 0.388 & 0.702  
          & 0.167 & 0.438  
          & 0.233 & 0.431 \\
    & BN Adapt \cite{wang2021tent} 
          & 0.775 & 3.465  
          & 2.928 & 8.209  
          & \third{0.494} & \third{1.921}  
          & & \third{0.147} & \third{0.271}  
          & \third{0.120} & \third{0.345}  
          & \third{0.082} & \third{0.181} \\
    & CoTTA \cite{wang2022continual} 
          & 0.767 & 3.799  
          & \third{2.650} & 6.242  
          & 0.933 & 2.763  
          & & 0.390 & 0.704  
          & 0.205 & 0.540  
          & 0.234 & 0.496 \\
    & TEA \cite{yuan2024tea} 
          & \third{0.735} & \third{3.417} 
          & 2.841 & 6.667 
          & 1.055 & 2.942 
          & & 0.256 & 0.411 
          & 0.206 & 0.415 
          & 0.132 & 0.226 \\
    & ProxyTTA \cite{park2024test} 
          & \second{0.686} & \second{2.666}  
          & \second{2.589} & \second{6.006}  
          & \second{0.477} & \second{1.598}  
          & & \second{0.124} & \second{0.240}  
          & \second{0.113} & \second{0.333}  
          & \second{0.074} & \second{0.166} \\
    & ETA (Ours)      
          & \best{0.545} & \best{2.617}  
          & \best{2.359} & \best{5.927}  
          & \best{0.472} & \best{1.568}  
          & & \best{0.105} & \best{0.204}  
          & \best{0.109} & \best{0.311}  
          & \best{0.067} & \best{0.154} \\
\midrule
\multirow{6}{*}{CostDCNet \cite{kam2022costdcnet}} 
    & Pre-trained     
          & 1.042 & 6.301  
          & 3.064 & 6.630  
          & 1.093 & 2.798  
          & & 0.189 & 0.446  
          & 0.173 & 0.443  
          & 0.144 & 0.458 \\
    & BN Adapt \cite{wang2021tent} 
          & \third{0.729} & \third{3.413}  
          & \third{2.288} & 6.338  
          & \third{0.469} & \second{1.572}  
          & & \third{0.136} & \third{0.338}  
          & \third{0.134} & \third{0.385}  
          & \third{0.068} & \third{0.164} \\
    & CoTTA \cite{wang2022continual} 
          & 0.756 & 3.686  
          & 2.676 & \third{6.09
          9}  
          & 0.689 & 2.140  
          & & 0.147 & 0.376  
          & 0.136 & 0.405  
          & 0.101 & 0.322 \\
    & TEA \cite{yuan2024tea} 
          & 0.786 & 3.697  
          & 2.841 & 6.667  
          & 0.738 & 1.748  
          & & 0.156 & 0.278   
          & 0.206 & 0.415  
          & 0.933 & 0.301 \\
    & ProxyTTA \cite{park2024test} 
          & \second{0.512} & \second{2.735}  
          & \second{2.062} & \second{5.509}  
          & \second{0.466} & \third{1.580}  
          & & \second{0.095} & \second{0.203}  
          & \second{0.125} & \best{0.357}  
          & \second{0.068} & \second{0.162} \\
    & ETA (Ours)      
          & \best{0.508} & \best{2.629}  
          & \best{2.048} & \best{5.439}  
          & \best{0.463} & \best{1.569}  
          & & \best{0.089} & \best{0.189}  
          & \best{0.117} & \second{0.364}  
          & \best{0.059} & \best{0.152} \\
\midrule
\multirow{6}{*}{BP-Net \cite{tang2024bpnet}} 
    & Pre-trained     
          & 0.893 & 3.926  
          & 2.787 & 6.500  
          & 0.692 & 2.534  
          & & 0.234 & 0.405  
          & 0.271 & 0.488  
          & 0.123 & 0.196 \\
    & BN Adapt \cite{wang2021tent} 
          & \third{0.605} & 3.426 
          & 2.618 & 5.850 
          & 0.540	& \third{1.838}
          & & 0.178 & 0.257
          & \third{0.236} & \third{0.373} 
          & 0.116 & 0.177 \\
    & CoTTA \cite{wang2022continual} 
          & 0.611 & \third{3.090} 
          & \third{2.432} & \third{5.490}  
          & \third{0.538} & 1.874
          & & 0.212 & 0.364  
          & 0.293 & 0.421 
          & 0.140 & 0.199 \\
    & TEA \cite{yuan2024tea} 
          &  0.631 & 	3.311		
          & 2.581 &  5.677
          & 0.673	 &  1.942
          & & \third{0.171} & \third{0.253} 
          & 0.256 & 0.404  
          & \third{0.115} & \third{0.165} \\
    & ProxyTTA \cite{park2024test} 
          & \second{0.571} & \second{2.844}
          & \second{2.373} & \second{5.413}
          & \second{0.489} & \second{1.520}
          & & \third{0.174} & \second{0.248}
          & \second{0.231} & \second{0.368}
          & \second{0.102} & \second{0.155} \\
    & ETA (Ours)      
          & \best{0.544} & \best{2.729}
          & \best{2.281} & \best{5.278}
          & \best{0.451} & \best{1.433} 
          & & \best{0.161} & \best{0.231}
          & \best{0.221} & \best{0.352}
          & \best{0.093} & \best{0.148} \\
\bottomrule
\end{tabular}%
}
\vspace{-3mm}
\caption{\textit{Quantitative results.} For outdoors, we adapt from KITTI to VKITTI-FOG, nuScenes, and Waymo; for indoors, from VOID to NYUv2, SceneNet, and ScanNet. We report MAE and RMSE in \textit{meters}. \highlight{tabfirst}{red} denotes best, \highlight{tabsecond}{orange}  second-best, and \highlight{tabthird}{yellow}  third-best.}
\vspace{-2.8mm}
\label{tab:experiments:all}
\end{table*}

\vspace{1pt}\noindent\textbf{Adaptation Objective.}
Finally, we integrate the three aforementioned loss components into a unified adaptation objective. Our adaptation loss is a linear combination:
\begin{equation}
\mathcal{L}_{\mathrm{adapt}} \;=\; w_{e}\,\ell_{\mathrm{energy}} \;+\; w_{z}\,\ell_{\mathrm{sparse}} \;+\; w_{s}\,\ell_{\mathrm{smooth}},
\label{eq:loss}
\end{equation}
where the weights $w$ balance the contributions of each loss term during test-time optimization. We will update the adaptation parameters $\psi$ by minimizing \cref{eq:loss} on each sequentially observed example or batch of testing data $D_t$.

\section{Experiments}
\label{sec:method}


\subsection{Experimental Setup}
\label{sec:exp_setup}
\vspace{1pt}\noindent\textbf{Datasets.}
We evaluate our method on a diverse range of datasets, spanning both real-world and synthetic domains. For adaptations in \textit{indoor} settings, we use: \textbf{VOID} \cite{wong2020unsupervised}, a real-world dataset captured using XIVO \cite{fei2019geo} (e.g., classrooms, laboratory, etc.); \textbf{SceneNet} \cite{mccormac2016scenenet}, a synthetic dataset with diverse room layouts and furniture arrangements; \textbf{ScanNet} \cite{dai2017scannet}, a large-scale indoor dataset captured with Structure Sensor; \textbf{NYUv2} \cite{Silberman:ECCV12}, providing household, office, and commercial environments captured with a Microsoft Kinect.
For \textit{outdoor} driving scenarios, we use: \textbf{KITTI} \cite{geiger2013vision}, a widely-used autonomous driving benchmark captured in daytime conditions using Velodyne LiDAR sensors; \textbf{VKITTI} \cite{gaidon2016virtual}, a synthetic dataset designed to replicate and augment KITTI; \textbf{Waymo} \cite{sun2020scalability}, providing large-scale autonomous driving data across diverse driving scenarios. Further details on the dataset are provided in Supp. Mat.

\vspace{1pt}\noindent\textbf{Models.} For evaluation, we employ ETA on four widely-used architectures for depth completion: MSG-CHN \cite{li2020multi}, NLSPN \cite{park2020non}, CostDCNet \cite{kam2022costdcnet} and BP-Net \cite{tang2024bpnet}. For each depth completion model, an energy-based model is trained on the same source dataset as the depth completion model. We evaluate performance using standard metrics in depth completion: Mean Absolute Error (MAE) and Root Mean Squared Error (RMSE), both reported in \textit{meters}. Exact equations of metrics are provided in Supp. Mat.


\vspace{1pt}\noindent\textbf{Baseline Methods.}
We compare ETA against four baseline methods for test-time adaptation: TENT \cite{wang2021tent}, CoTTA \cite{wang2022continual}, TEA \cite{yuan2024tea}, and ProxyTTA \cite{park2024test}. Since the original implementations of TENT and TEA are not directly applicable to our task, we employ adapted versions. Specifically, \textit{BN Adapt} denotes a variant of TENT, which minimizes \cref{eq:sparse,eq:smooth} instead of entropy. Likewise for TEA, we interpret the sum of \cref{eq:sparse,eq:smooth} as ``energy'', since the log-sum-exp of classifier logits cannot be computed in depth completion tasks. 
Further details can be found in Supp. Mat.

\subsection{Main Results}
\label{sec:results}
We report the performance of ETA in both \textit{outdoor} and \
\textit{indoor} settings. For each scenario, the models are pre-trained on KITTI \cite{geiger2013vision} and VOID \cite{wong2020unsupervised}, respectively. 

\vspace{1pt}\noindent\textbf{Main Result.} As shown in~\cref{tab:experiments:all}, ETA consistently achieves superior performance, significantly lowering both Mean Absolute Error (MAE) and Root Mean Squared Error (RMSE) across all tested domains. In the outdoor setting (e.g.,VKITTI-FOG, nuScenes, and Waymo), ETA surpasses baseline methods, including the previous state-of-the-art ProxyTTA \cite{park2024test} by an average of 5.36\% and 1.97\% in MAE and RMSE, respectively. ProxyTTA specifically targets depth completion by learning a proxy embedding from the sparse depth modality, which is less sensitive to domain shifts, to guide adaptation of image features towards source-domain distributions.
However, it uniformly adapts the adaptation layer parameters across the entire predicted depth map, and solely relies on the proxy embedding from sparse depth inputs. This does not account for local errors in the depth map prediction and can potentially mislead the adaptation of the regions not covered by sparse depth. In contrast, the proposed energy model predicts the likelihood of the depth predictions from dense depth map prediction conditioned on the sparse point cloud as an energy map. Hence, even on the regions missing sparse depth, the energy model can still quantify energy.

We also compare our method against TEA \cite{yuan2024tea}, which is the closest existing work to our method, using an energy-based objective originally designed for classification tasks.
TEA does not explicitly address continuous regression tasks or localized correction of errors.
ETA gains notable average improvement of 24.9\% and 19.0\% on MAE and RMSE against TEA.
Similar improvements are observed on nuScenes and Waymo, underscoring ETA's adaptation ability to challenging outdoor conditions such as fog and varying illumination.
The improvement comes from ETA is even more prominent when adapting models that initially exhibit a larger errors after being transferred to the target testing data (e.g., KITTI$\rightarrow$VKITTI-FOG on MSG-CHN). This demonstrates ETA's ability to align source and target distributions, and enable less performant models to produce high-fidelity outputs in the target domain. The latest depth completion models, such as BP-Net, see smaller but still consistent improvements, which aligns with our expectation that the energy-based model identifies fewer erroneous regions when the initial error is already low.

\begin{figure*}[t]
    \centering
    \includegraphics[width=1.0\linewidth]{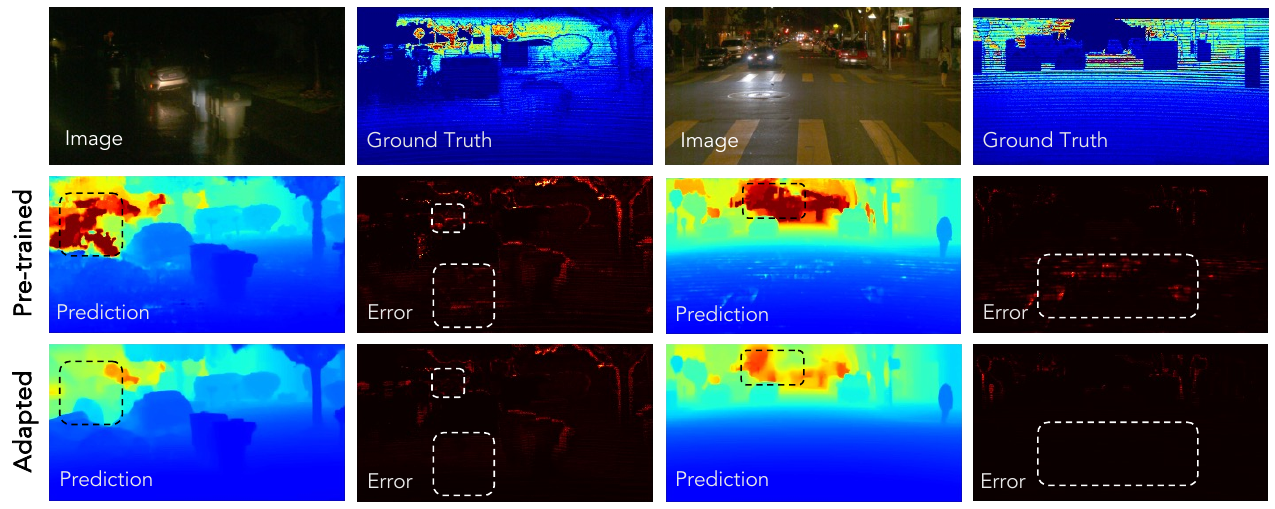}
    \vspace{-6mm}
    \caption{\textit{Qualitative results on Waymo-Hard.} We adapt CostDCNet from KITTI$\rightarrow$Waymo-Hard. ETA can effectively adapt to adverse weather conditions. Pre-trained denotes predictions using KITTI trained model, and adapted denotes results after TTA.
    }
    \label{fig:adv_weather_visualization}
    \vspace{-3mm}
\end{figure*}

In indoor environments, including NYUv2, SceneNet, and ScanNet, ETA also demonstrates clear advantages, achieving an average of 10.13\% and 10.33\% improvement over the previous state-of-the-art in MAE and RMSE. Consistent with results from outdoors, we again improve by and 58.2\% and 27.13\% over TEA. These results demonstrate that our method can successfully guide model adaptation when transferred to a novel testing data distribution. 

\vspace{1pt}\noindent\textbf{Generalization to Adverse Conditions.}
We further assess the robustness of ETA on a challenging subset of the Waymo dataset, which we denote as \textit{Waymo-Hard}. It specifically contains night-time scenes, adverse weather, and highly dynamic scenarios with blurred moving vehicles and pedestrians. As shown in~\cref{tab:adverse_weather} and~\cref{fig:adv_weather_visualization}, ETA outperforms ProxyTTA in both MAE and RMSE, effectively bridging the domain gap by severe adverse conditions.

\vspace{1pt}\noindent\textbf{Generalization Across Large Domain Shifts.} ETA is evaluated on a particularly challenging domain shift that we designed: Adapting a model from an outdoor driving environment to indoor scenes. Specifically, BP-Net model trained on KITTI is adapted to the NYUv2, SceneNet, and ScanNet datasets.
As shown in \cref{tab:outdoor_to_indoor}, the baseline methods suffer from a significant performance degradation coming from the large gap in the prediction range and the appearance of the 3D scene. Nevertheless, ETA consistently achieves the best performance across all three indoor target domains, demonstrating its robustness even when the domain discrepancy is larger than the previously tested (e.g., indoor-to-indoor and outdoor-to-outdoor).

\begin{table}[t]
\small 
\centering
\begin{adjustbox}{width=1.0\linewidth}
\setlength\tabcolsep{2pt}
\begin{tabular}{clcc}
    \toprule
    Method & & MAE & RMSE \\
    \midrule 
    \multirow{3}{*}{\,\,\,\,\,\,CostDCNet \cite{kam2022costdcnet}\,\,\,\,\,} & \,\,\,\,\,\,Pre-trained\,\,\,\,\,\, & 0.776 & 2.072 \\
    & \,\,\,\,\,\,ProxyTTA\,\,\,\,\,\,
    & 0.595 & 1.728 \\ 
    & \,\,\,\,\,\,ETA (Ours)\,\,\,\,\,\, & \,\,\,\,\,\,\best{0.582}\,\,\,\,\,\, & \,\,\,\,\,\,\best{1.699}\,\,\,\,\,\, \\
    \bottomrule
\end{tabular}
\end{adjustbox}
\vspace{-1mm}
\caption{\textit{Quantitative results on Waymo-Hard.} We adapt CostDCNet from KITTI $\rightarrow$ Waymo-Hard. The best result in \highlight{tabfirst}{red}.}
\vspace{-3mm}
\label{tab:adverse_weather}
\end{table}

\begin{table}[t]
\small 
\centering
\begin{adjustbox}{width=1.0\linewidth}
\setlength\tabcolsep{2pt}
\begin{tabular}{clcc}
    \toprule
    Method & & MAE & RMSE \\
    \midrule 
    \multirow{3}{*}{\,\,\,\,\,\,MSG-CHN \cite{li2020multi}\,\,\,\,\,} & \,\,\,\,\,\,Pre-trained\,\,\,\,\,\, & 2.842 & 6.557 \\
    & \,\,\,\,\,\,Global (Image-based)\,\,\,\,\,\,
    & 1.406 & 4.226 \\ 
    & \,\,\,\,\,\,Local (Region-based)\,\,\,\,\,\, & \,\,\,\,\,\,\best{0.703}\,\,\,\,\,\, & \,\,\,\,\,\,\best{2.996}\,\,\,\,\,\, \\
    \bottomrule
\end{tabular}
\end{adjustbox}
\vspace{-2mm}
\caption{\textit{Comparison of different ETA update strategies.} We adapt MSG-CHN from KITTI $\rightarrow$ VKITTI-FOG. The best result in \highlight{tabfirst}{red}.}
\vspace{-4mm}
\label{tab:eta_comparison}
\end{table}


\subsection{Ablation Studies}
\label{sec:ablations}

\begin{figure*}[t!]
    \centering
    \resizebox{1.0\linewidth}{!}{ 
        \includegraphics{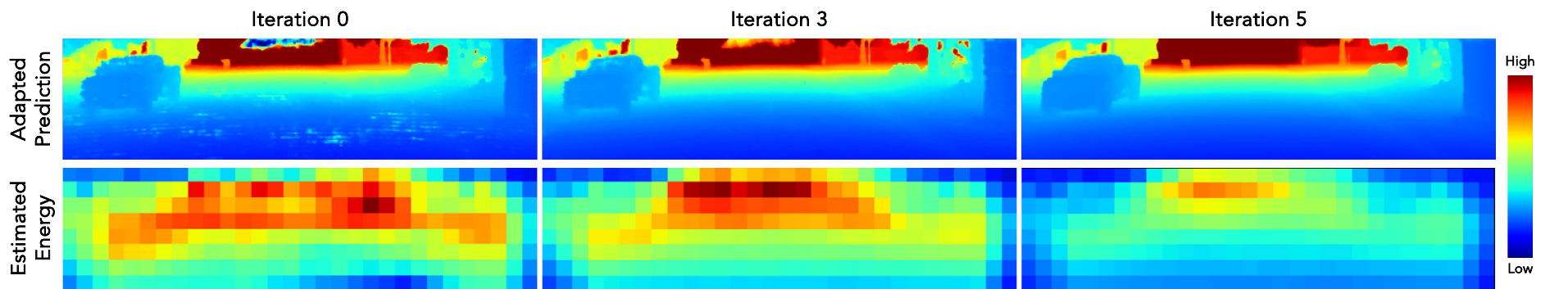}
    }
    \vspace{-6mm}
    \caption{\textit{Qualitative evaluation of ETA across iterations.} We adapt from KITTI $\to$ Waymo. The initial predictions were made with the pre-trained model. The estimated energy effectively highlights the regions to be corrected, guiding the model to make localized changes.}
    \label{fig:iterations}
    \vspace{-6mm}
\end{figure*}

\begin{figure}[th!]
    \centering
    \includegraphics[width=0.92\linewidth]{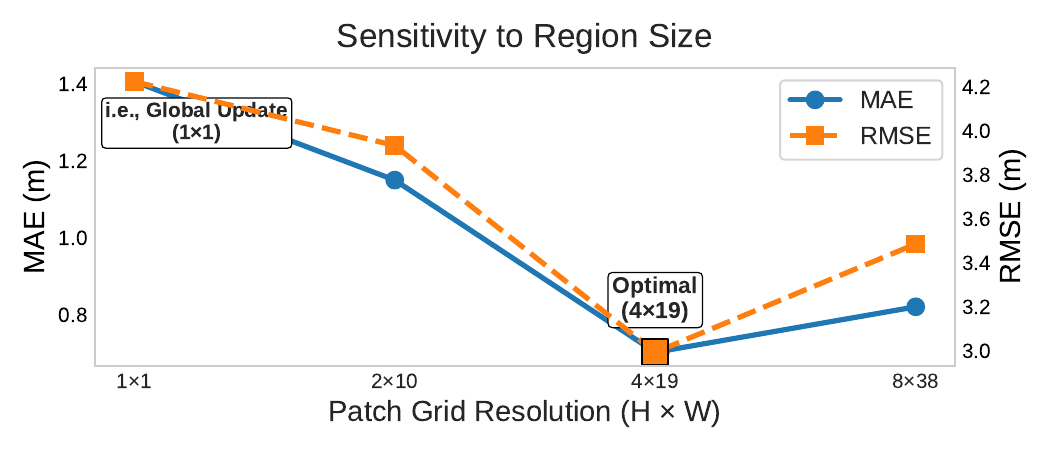}
    \vspace{-2mm}
    \caption{\textit{Sensitivity to output energy map size.} We adapt MSG-CHN from KITTI $\to$ VKITTI-FOG. Smaller patch sizes (e.g., 1$\times$1) correspond to global updates, and larger sizes (e.g., 8$\times$38) imply highly localized updates.}
    \label{fig:region_size_ablation}
    \vspace{-4mm}
\end{figure}

\vspace{1pt}\noindent\textbf{Energy Model with Varying Region size.}
Energy values allow us to make localized adjustments, and the area affected by the adjustment depends on the fidelity of energy predictions.
An extreme case is to have a single energy value accounting the errors in the entire prediction, which ambiguates the regions with error.
Our region-based energy formulation mitigates this issue by providing updates by minimizing the energy with areas in the granular level, as demonstrated in~\cref{fig:region_size_ablation}.
Further, this region-based energy model provides a coarse level of interpretability in the adaptation process by highlighting the contribution of each prediction region in energy, shown in~\cref{fig:iterations}.


\begin{table}[t]
  \centering
  \small
  \setlength{\tabcolsep}{3pt}
  \begin{adjustbox}{width=\columnwidth,center}
    \begin{tabular}{l l c c c c c c}
      \toprule
      & & \multicolumn{2}{c}{NYUv2} & \multicolumn{2}{c}{SceneNet} & \multicolumn{2}{c}{ScanNet} \\
      \cmidrule(lr){3-4} \cmidrule(lr){5-6} \cmidrule(lr){7-8}
      Model      & Method       & MAE$\downarrow$ & RMSE$\downarrow$ & MAE$\downarrow$ & RMSE$\downarrow$ & MAE$\downarrow$ & RMSE$\downarrow$ \\
      \midrule
      \multirow{6}{*}{BP-Net \cite{tang2024bpnet}}
                 & Pre-trained  & 1.987           & 2.578            & 1.432           & 1.872            & 2.657           & 3.416            \\
                 & BN Adapt     & 1.343           & 1.839            & 0.423           & 0.634            & 0.325           & 0.480            \\
                 & CoTTA        & 1.401           & 1.881            & 0.411           & 0.591            & 0.381           & 0.515            \\
                 & TEA          & 1.392           & 1.856            & 0.409           & 0.617            & 0.348           & 0.501            \\
                 & ProxyTTA     & 1.380           & 1.874            & 0.401           & 0.583            & 0.311           & 0.464            \\
                 & ETA (Ours)   & \best{1.322}  & \best{1.807}   & \best{0.340}  & \best{0.521}   & \best{0.272}  & \best{0.406}   \\
      \bottomrule
    \end{tabular}
  \end{adjustbox}
  \vspace{-2mm}
  \caption{\textit{Quantitative results on outdoor-to-indoor adaptation.} We adapt BP-Net from KITTI to NYUv2, SceneNet and ScanNet. ETA consistently achieves the best results (\highlight{tabfirst}{red}).}
  \label{tab:outdoor_to_indoor}
  \vspace{-4mm}
\end{table}

\begin{figure}[t]
    \centering
    \includegraphics[width=0.92\linewidth]{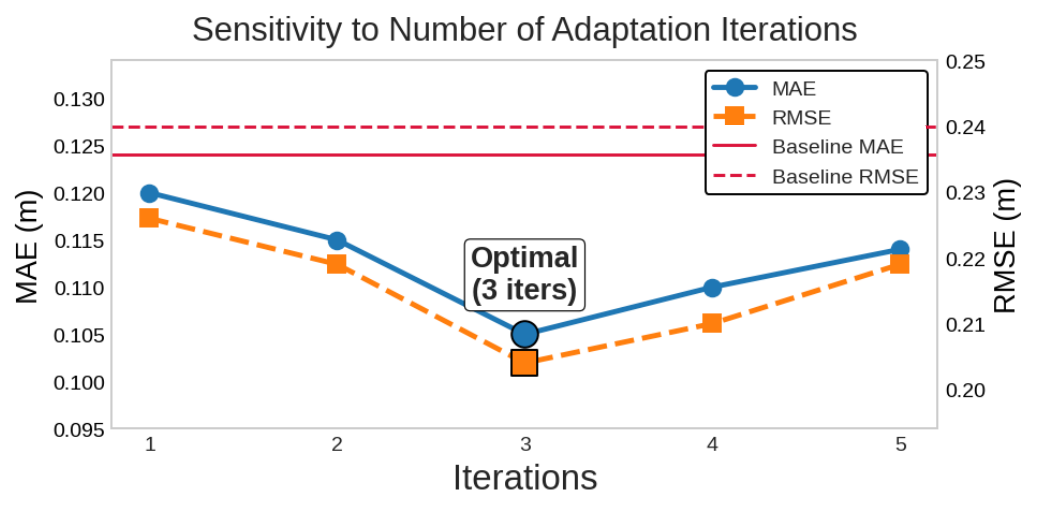}
    \vspace{-2mm}
    \caption{\textit{Sensitivity to Adaptation Iterations.} We adapt NLSPN from VOID $\to$ NYUv2. Solid and dashed red lines represent baseline MAE and RMSE, respectively.
    }
    \label{fig:iteration_sensitivity}
    \vspace{-3mm}
\end{figure}

\vspace{1pt}\noindent\textbf{Sensitivity on Iterations.} Increasing the number of adaptation iterations adds the computational budget but may offer better adaptation result.
\cref{fig:iteration_sensitivity} demonstrates this sensitivity for NLSPN adapting from VOID to NYUv2. While optimal performance is achieved at three iterations in this case, we note that in most scenarios, a single iteration—the lower bound—is sufficient to attain state-of-the-art results.

\begin{figure}[t]
    \centering
    \includegraphics[width=0.92\linewidth]{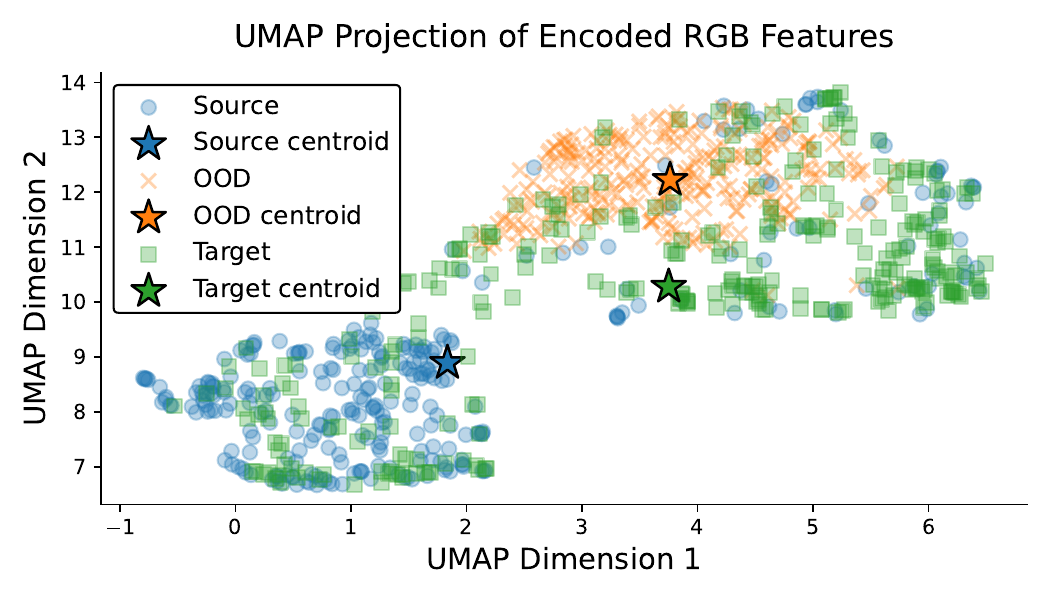}
    \vspace{-4mm}
    \caption{\textit{UMAP plot of RGB features.} Adversarially perturbed examples from the source dataset (VOID) closely approximate the target domain (NYUv2), to bridge the domain gap.}
    \label{fig:umap}
    \vspace{-5mm}
\end{figure}

\vspace{1pt}\noindent\textbf{Generation Quality.} We assess the quality of our synthetic out-of-distribution (OOD) samples in modeling domain gaps. For visualization, we adversarially perturb source data (VOID) to synthesize target (NYUv2) samples. The UMAP plot in~\cref{fig:umap} places synthetic OOD samples closer to target (NYUv2) than source (VOID) clusters, showing that our synthetic OOD data can effectively serve as a proxy for the target distribution.


\section{Discussion}
\label{sec:discussion}

Our experiments show that ETA achieves superior adaptation performance in diverse indoor and outdoor scenarios, consistently outperforming test-time adaptation baselines by a noticeable margin.
We attribute these gains to three main factors. First, the energy-based model offers a flexible way to quantify how well the predictions align with the training distribution. Second, the region-based formulation localizes errors and focuses on specific regions in need of correction. Third, restricting updates to a lightweight adaptation layer allows the rest of the network to remain stable, mitigating the risk of catastrophic forgetting that can arise during iterative test-time updates.

Notably, ETA demonstrates strong robustness in challenging environments such as foggy or nighttime outdoor scenes, and highly cluttered indoor scenarios. These conditions often impose large domain shifts that degrade depth completion accuracy. By detecting out-of-distribution regions and adjusting only a small set of parameters, our approach consistently brings improvement in both quantitative metrics (MAE, RMSE) and qualitative visualizations. Beyond depth completion, the energy-based test-time adaptation principle can be extended to other dense regression tasks like optical flow, surface normal prediction, or medical image segmentation, where continuous outputs must be adapted to various lighting, weather, and sensor conditions.

\vspace{1pt}\noindent\textbf{Limitations.} Although our method demonstrates notable improvements, certain limitations remain. First, our region energy approach avoids large-scale, uniform corrections across the entire image, yet some domain shifts may require broader scene-level updates. If the target environment exhibits drastically different global characteristics, the localized corrections alone may be insufficient to account for large variations in scale or depth ranges. Second, our energy-based model relies on adversarial perturbations to approximate out-of-distribution samples in the source domain. This design may be less effective if the real target domain features extreme conditions, sensor anomalies, or semantic shifts not captured by the perturbations.


\paragraph{Acknowledgments}
This work is supported by the NSF
2112562 Athena AI Institute, NIH/NHLBI R01HL121226, IITP-RS-2021-II211341 and NRF-RS-2023-00251366.

{
    \small
    \bibliographystyle{ieeenat_fullname}
    \bibliography{main,lab,adversarial_perturbations}
}

\newpage

\appendix

\twocolumn[{
 \centering
  {\Large{\textbf{ 
  ETA: Energy-based Test-time Adaptation for Depth Completion \vspace{0.5cm}
  \\ SUPPLEMENTARY MATERIAL}}}
  \vspace{1cm}
}]

\section{Datasets}
\label{sec:datasets}

\textbf{KITTI} dataset \cite{geiger2013vision} provides calibrated RGB images synchronized with Velodyne lidar point clouds, GPS, and inertial data, collected from over 61 driving scenes. It includes $\approx$80K raw image frames paired with sparse depth maps of $\approx$5\% density, commonly used for depth completion \cite{uhrig2017sparsity}. Semi-dense depth data is available for the bottom 30\% of the image space, while ground-truth depth maps combine 11 consecutive raw lidar scans. We trained our model on $\approx$86K single images, without using the test or validation sets.

\noindent\textbf{VOID} dataset \cite{wong2020unsupervised} consists of 640$\times$480 RGB images synchronized with sparse depth maps captured in indoor settings like classrooms and laboratories, and outdoor gardens. Sparse depth maps ($\approx$0.5\% density, $\approx$1,500 points) were created with the XIVO VIO system \cite{fei2019geo}, while dense ground-truth maps were obtained using active stereo. VOID introduces challenging 6 DoF motion due to rolling shutter effects in 56 sequences, contrasting with KITTI's planar motion. Our model was trained on $\approx$46K images.

\noindent\textbf{NYUv2} dataset \cite{Silberman:ECCV12} contains 372K synchronized 640$\times$480 RGB images and depth maps captured using Microsoft Kinect across 464 indoor scenes, including homes, offices, and stores. To simulate SLAM/VIO-style sparse depth maps, we employed the Harris corner detector \cite{harris1988combined} to extract $\approx$1,500 points from the depth maps. We evaluated adaptation performance on 654 test images.

\noindent\textbf{ScanNet} dataset \cite{dai2017scannet} offers 2.5 million images with dense depth maps across 1,513 indoor scenes. SLAM/VIO-style sparse depth maps were simulated by applying the Harris corner detector \cite{harris1988combined}, sampling $\approx$1,500 points from the dense maps. Our experiments utilized $\approx$21K test images for adaptation.

\noindent\textbf{Virtual KITTI (VKITTI)} dataset \cite{gaidon2016virtual} includes $\approx$17K 1242$\times$375 synthetic images across 35 videos, derived from 5 original KITTI videos augmented with 7 variations in lighting, weather, and camera perspectives \cite{uhrig2017sparsity}. To minimize the large domain gap between RGB images from VKITTI and KITTI despite Unity's virtual similarity to KITTI scenes \cite{gaidon2016virtual}, we used VKITTI's dense depth maps only to reduce the domain gap in photometric variations, while sparse depth maps were simulated to match KITTI's lidar-generated distribution in terms of marginal distribution of sparse points. A test set of $\approx$2,300 images was used for adaptation.

\noindent\textbf{nuScenes} dataset \cite{caesar2020nuscenes} provides 1600$\times$900 RGB images synchronized with sparse point clouds, featuring 27.4K training images from 1,000 driving scenes and 5.8K test images from 150 scenes. For the test set, ground truth was created by merging projected sparse depth from forward-backward frames. Setup details will be provided with released code for reproducibility.

\noindent\textbf{SceneNet} dataset \cite{mccormac2016scenenet} comprises 5 million 320$\times$240 RGB images with depth maps captured in simulated indoor environments with randomized room arrangements. Due to the lack of sparse depths, sparse depth maps were derived using the Harris corner detector \cite{harris1988combined} simulating SLAM/VIO outputs, followed by k-means clustering to reduce the sampled points to 375 (0.49\% total pixel density). We used $\approx$2,300 test images for adaptation from a single split (out of 17 available) of 1,000 sequences of 300 images each. Each sequence is generated by recording the same scene over a trajectory.

\noindent\textbf{Waymo Open Dataset} \cite{sun2020scalability} includes 1920$\times$1280 RGB images and lidar scans collected at 10Hz in autonomous vehicle scenes. It features $\approx$158K training images from 798 scenes, and $\approx$40K validation images from 202 scenes with sampling frquency of 0.6 seconds. Objects are annotated across full 360$^{\circ}$ field. Each top lidar sensor's point cloud is projected onto camera frame. Ground truth was generated by merging top and front lidar scans projected over 10 forward-backward frames, corresponding to 1-second intervals, with moving objects removed using annotations. Outliers in depth points were filtered out for accuracy.

\begin{table}[t]
             \centering 
             \footnotesize
             \renewcommand{\arraystretch}{1.2}
             \begin{tabular}{cccccc} \hline 
                  Dataset &  LR &  $w_{sm}$&  $w_{z}$&  $w_{energy}$ & Inner Iter.\\\hline\hline
 \multicolumn{6}{c}{MSG-CHN}\\\hline
                  Waymo&  3e-3&  3.0&  1.0&  0.001& 3\\ 
                  VKITTI-FOG&  5e-4&  6.0 & 1.0 & 0.5 & 5\\  
        nuScenes&  3e-3&  5.0&  1.0&  0.5& 3\\ 
                  SceneNet&  1e-3&  8.0&  1.0&  0.1& 3\\
                  NYUv2&  5e-4&  7.5&  1.0&  0.004& 3\\ 
                  ScanNet&  5e-3&  8.0&  1.0&  0.001& 3\\\hline
 \multicolumn{6}{c}{NLSPN}\\\hline
 Waymo& 6e-3& 1.0& 1.0& 0.001 &1\\
 VKITTI-FOG& 1e-3& 1.0& 1.0& 0.001&1\\
 nuScenes& 6e-3& 1.0& 1.0& 0.002&1\\
 SceneNet& 3e-3& 1.5& 1.0& 2.0&3\\
 NYUv2& 4e-3& 5.0& 1.0& 1.0&3\\
 ScanNet& 1e-4& 2.0& 1.0& 0.3&3\\ \hline 
                  \multicolumn{6}{c}{CostDCNet}\\\hline
 Waymo& 5e-3& 3.0 & 1.0& 0.1&1\\
 VKITTI-FOG& 5e-3& 3.0& 1.0& 0.04&1\\  
                  nuScenes&  5e-3&  3.0&  1.0&  0.003& 1\\ 
                  SceneNet&  6e-3&  2.5&  1.0&  0.001& 3\\ 
                  NYUv2&  3e-3&  3.5&  1.0&  0.0001& 3\\ 
 ScanNet& 2e-3& 2.0& 1.0& 0.0002&3\\\hline
             \end{tabular}
             \caption{\textit{Hyperparameters.} Model specific hyperparameters used at test-time.}
             \label{tab:hyperparameter}
\vspace{-5mm}
\end{table}

\begin{figure*}[t!]
    \centering
    \includegraphics[width=1.0\linewidth]{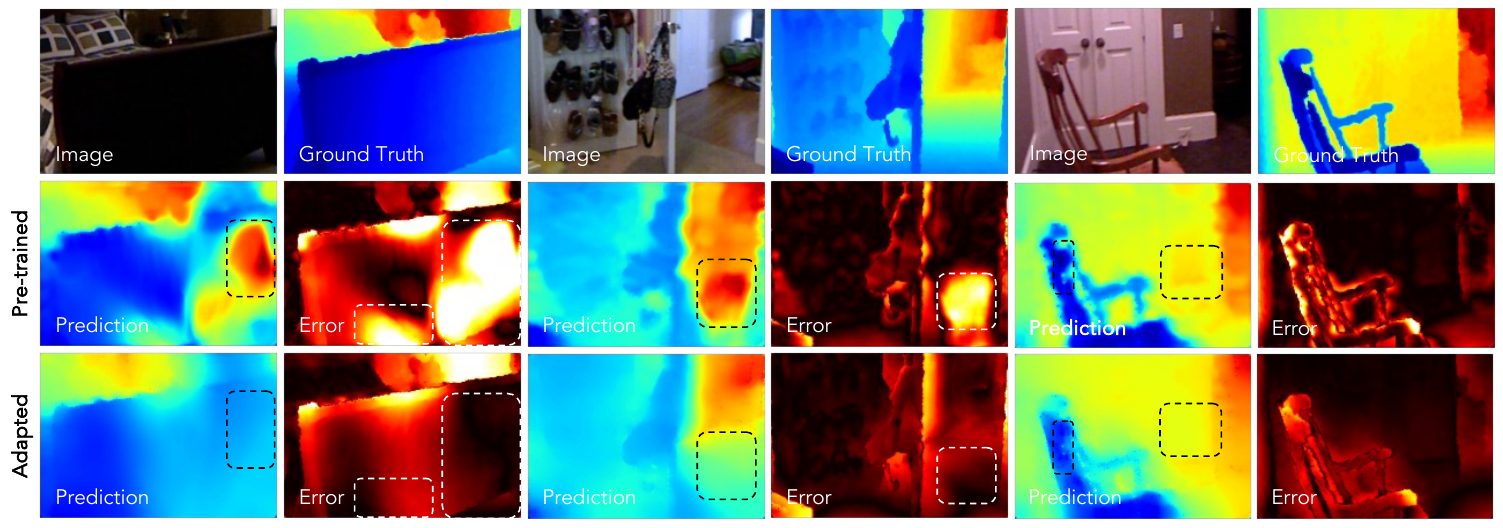}
    \vspace{-6mm}
    \caption{\textit{Qualitative results on NYUv2.} We adapt CostDCNet from VOID$\rightarrow$NYUv2.
    }
    \label{fig:voidtonyuv2}
    \vspace{-4mm}
\end{figure*}

\section{Implementation and training details}
\noindent\textbf{Model Architecture.} Energy model is implemented as a convolutional neural network that takes a two-channel input of sparse depth and the dense prediction. It uses six 5x5 convolutional layers (stride 2) with LeakyReLU activations to increase channel depth from 2 to 512. A final 3x3 convolutional layer then maps these features to a single-channel energy map to score input regions. \\
\noindent\textbf{Hyperparameters.} Model and dataset specific hyperparameters for test-time adaptation are noted in \ref{tab:hyperparameter}. \\
\noindent\textbf{Training energy models.} We take baseline depth completion models pre-trained on KITTI and VOID from~\cite{park2024test}. For each model, we train patch-based energy model on the corresponding source dataset \ie KITTI, VOID. All models were trained for 5 epochs with a batch size of 32. Specific learning rates and hyperparameters for data augmentation will be released with the code.\\
\noindent\textbf{Evaluation.} For outdoor datasets, test-time adaptation performances are evaluated on bottom-cropped regions to exclude regions where no corresponding sparse depth exists. For VKITTI, we evaluate on $1240\times240$ bottom-cropped regions, $1600\times544$ for nuScenes, and $1920\times640$ for Waymo. For indoor datasets, models are evaluated on the entire region. The error metrics used for evaluation are defined in \ref{tab:error_metrics}. For outdoor, we evaluate the models on depth range from 0.0 to 80.0 meters. For indoor, we evaluate on 0.2 to 5.0 meters.

\begin{table}[t]
\centering
\footnotesize
\setlength\tabcolsep{20pt}
\begin{tabular}{l l}
    \midrule
        Metric & Definition \\ \midrule
        MAE &$\frac{1}{|\Omega|} \sum_{x\in\Omega} |\hat d(x) - d_{gt}(x)|$ \\
        RMSE & $\big(\frac{1}{|\Omega|}\sum_{x\in\Omega}|\hat d(x) - d_{gt}(x)|^2 \big)^{1/2}$ \\
        \midrule
    \end{tabular}
    \caption{
        \textit{Error metrics.} $d_{gt}$ means the ground-truth depth.
    }
    \vspace{-1.6em}
\label{tab:error_metrics}
\end{table}

\section{Extended Related Work}
As we utilize adversarial perturbations in our method, we present a related works on the topic as an extended discussion.

\noindent\textbf{Adversarial Perturbations.} Small input perturbations can significantly alter classification outputs \cite{szegedy2013intriguing}. Goodfellow et al. \cite{goodfellow2014explaining} introduced Fast Gradient Sign Method (FGSM), later extended to iterative variants for increased effectiveness \cite{madry2017towards,dong2018boosting}. Minimal perturbations were studied in \cite{moosavi2016deepfool}, and lower bounds on their magnitudes were analyzed in \cite{peck2017lower}. Adversarial examples can yield high-confidence outputs from unrecognizable inputs \cite{nguyen2015deep}, and are attributed to non-robust features \cite{ilyas2019adversarial}. Transferability across models and datasets was explored in \cite{xie2019improving,naseer2019cross}.



Adversarial attacks has also been studied in dense prediction tasks.
Prior works considered detection and segmentation \cite{xie2017adversarial,hendrik2017universal}, monocular depth \cite{mopuri2018generalizable,wong2020targeted}, and optical flow \cite{ranjan2019attacking,schrodi2021causes}. 
Stereo attacks were considered in \cite{wong2021stereopagnosia}, and \cite{berger2022stereoscopic} studies universal perturbations for stereo depth estimation.
We exploit the adversarial perturbations as a mean of exploring the data space, where the perturbed samples simulates the out-of-distribution samples with source data. The out-of-distribution samples enable the energy model to learn to assign high energy to the predictions on target distribution.

\section{Additional Qualitative Results}
In addition to the qualitative results on the outdoor scenario presented in the main paper, Fig. \ref{fig:voidtonyuv2} shows the adaptation result of the pretrained CostDCNet~\cite{kam2022costdcnet} on the indoor scenario (VOID to NYUv2).
ETA demonstrates superior performance on homogeneous surfaces (left and middle columns) and boundary regions with texture discontinuities at similar depths (right column), as observed by the overall darker regions in the error maps. Boxes highlight the detailed comparisons.
These results indicate that ETA effectively adapts the depth completion model to previously unseen indoor environments characterized by background clutter, varying illumination, and novel objects, by minimizing energy learned from the in-distribution ``source'' data with and the simulated ``target'' data by adversarial perturbations.


\end{document}